\documentclass{article}

\usepackage{graphicx} 
\usepackage{subfigure}

\usepackage{etoolbox}
\usepackage{varioref}
\patchcmd{\abstract}
  {\bfseries\abstractname}
  {\scshape\abstractname}
  {}{}

\usepackage{algorithm}
\usepackage{algorithmic}
\PassOptionsToPackage{hyphens}{url}\usepackage{hyperref}
\usepackage{hyperref}
\usepackage{float}
\usepackage{datetime}

\usepackage[accepted]{icml2015}
\usepackage[superscript,biblabel]{cite}

\usepackage[accepted]{icml2015}

\newcommand{\timestamp}{compiled on \ifnum\day<10 0\fi\the\day.\,%
\ifnum\month<10 0\fi\the\month.\,%
\the\year\ at \xxivtime\,h}

\makeatletter
\def\namedlabel#1#2{\begingroup
   \def\@currentlabel{#2}%
   \label{#1}\endgroup
}
\makeatother

\def\thesection{\Roman{section}}                             
\def\thesubsection{\mbox{\Alph{subsection}.}}     

\icmltitlerunning{Thinking Required}

\usepackage[symbol]{footmisc}

\usepackage{titlesec}
\titleformat{\section}
  {\normalfont\scshape\centering}{\thesection}{1em}{}
  
\begin{document} 

\twocolumn[
\icmltitle{Thinking Required}

\icmlauthor{Kamil Rocki}{kmrocki@us.ibm.com}
\icmladdress{IBM Research, San Jose, CA 95120, USA}
\icmlkeywords{}

\vskip 0.3in
]

\begin{abstract}
There exists a theory of a single general-purpose learning algorithm which could explain the principles its operation. It assumes the initial rough architecture, a small library of simple innate circuits which are prewired at birth. and proposes that all significant mental algorithms are learned. Given current understanding and observations, this paper reviews and lists the ingredients of such an algorithm from architectural and functional perspectives.

\end{abstract}

\section{Background}
\label{intro}
In a very simplistic way, current research efforts in the field of Artificial Intelligence (AI) can be divided into\footnote{this might be the author's opinion only}: 
\begin{itemize}
\item \emph{classical AI}, also known as Good Old-Fashioned Artificial Intelligence\cite{Haugeland:1985:AIV:4694} (GOFAI), concerning well defined tasks and including optimization, rule-based inference, efficient knowledge representation, expert systems, planning, graph search algorithms\cite{Nilsson:80, Russell:2003:AIM:773294} and evolutionary algorithms\cite{opac-b1089183,simon2013evolutionary}{}, among others. 
\item \emph{statistical learning} (machine learning\cite{hastie01statisticallearning, Mitchell97a} or ML), where the data distribution is being fitted to a model describing the prior knowledge about a problem. It includes, but is not limited to, deep learning algorithms\cite{888, LeCun2015, Goodfellow-et-al-2015-Book}{}, reinforcement learning\cite{Watkins:1989, Tesauro:1995:TDL:203330.203343, Kaelbling96reinforcementlearning, Sutton98reinforcementlearning, DBLP:books/daglib/0035301, Mnih2015}{}, as well non-neural machine learning methods such as Support Vector Machines\cite{Vapnik:95} (SVMs). Despite not being fundamentally novel (working multi-layered networks already existed over 50 years ago\cite{ivakhnenko1965, ivakhnenko1971, Fukushima:1979neocognitron}{}), deep models have become very successful only recently, solving a variety of hard problems\cite{HinSal06, liwicki:icdar2007, mohamed2009, Graves:09tpami, ciresan:2010deepbig_arxiv, transfer2011, Ciresan:2012a, zeiler2013, NIPS2012_4824, Le_buildinghigh-level, coates:2013icml, Szegedy2013, graves2014, goodfellow2014multi, 1409.4842, DBLP:journals/corr/XuBKCCSZB15, russakovsky2014imagenet, karpathy2014}{}. The sudden popularity is due to their computational tractability, creation of large labeled data sets (e.g. ImageNet\cite{imagenet_cvpr09}{}), emergence of fast, programmable hardware for parallel processing\cite{buck, gpu2004, Harris:2005:FFD:1198555.1198790, Volkov:2008:BGT:1413370.1413402, chellapilla:2006b, raina2009large, uetz:2009, ciresan:2010deepbig_arxiv, ciresan:2011ijcai, ciresan:2012NN, coates:2013icml, 37631, farabet2011large}{}, usage of alternative nonlinearities such as rectifiers\cite{AISTATS2011_GlorotBB11, Dahl2013, Hinton_rectifiedlinear, maas2013rectifier, DBLP:journals/corr/HeZR015} and advances in numerical optimization \cite{DBLP:conf/icml/Martens10, LeNCLPN11, Martens:2011hessfree, NIPS2014_5486}{}. 
\item \emph{Artificial General Intelligence} (AGI) is  another branch of research which can be distinguished\cite{Hutter:02fast, Hutter:03unipriors, Hutter:04uaibook, quant-ph/0011122, Schmidhuber:02colt, Schmidhuber:02ijfcs, Hawkins:2004:INT:993636, creativity, kurzweil2012create} with ambitious theoretical description of what intelligence might be, however lacking practical algorithms which exist in machine learning. 
\item \emph{neuroscientific approach} is the forth one, concerned with building large, as anatomically accurate as possible, computational models\cite{koch1998methods, hawkins2015neurons, 10.3389/conf.fncom.2012.55.00033, seung2012connectome}{} in order to improve our understanding about the way brain works from the bottom-up way. The results of such research can be inspirational for the machine learning community (although recently, the deep learning algorithms give neuroscientists new about hierarchical representations in the brain).
\item \namedlabel{itm:neuromorphic}{\emph{neuromorphic}} \emph{neuromorphic},  brain-inspired, algorithm-agnostic hardware\cite{markram2012, Merolla08082014, journals/tc/FurberLPGPTB13, schemmeliscas2010, indiveri2011neuromorphic, lichtsteiner2008128, rachmuth2011biophysically}{}, focusing on modeling the physical properties of neurons and synapses in order to replicate brain's learning ability.
\end{itemize}

Recently, much progress has been made in the area of supervised learning (as it has been mentioned in the paragraph above). However, one of the greatest challenges remaining in artificial intelligence research is to make steps towards advancements in the field of unsupervised learning algorithms\cite{DBLP:journals/corr/abs-1206-5538, Bengio-2009, lecuncvpr, Goodfellow-et-al-2015-Book}{}. Especially, autonomous learning of complex spatiotemporal patterns (another way of thinking about predicting spatiotemporal patterns is structured prediction\cite{Bakir:2007:PSD:1296180}). The main motivation of the work presented in this paper is the observation that all human experiences are inherently spatiotemporal and that all predictions are context-dependent.

This paper postulates a need for another intensified research effort spanning the fields of neuroscience, machine learning, AGI and neuromorphic computing in order to design algorithms and build machines that think that is \emph{machine intelligence}. It focuses on the neocortex, that is the \emph{intelligence} part, not addressing other aspects related to the brain such as consciousness or emotions, reinforcement and long-term memory.





\section{Ingredients}



\subsection{General Purpose}
Neocortex, which is attributed only to mammals, is deemed to be the place where intelligent behavior resides. It has been studied extensively over the past decades, but to date there is still no consensus on how it works. There exists a theory of a single learning algorithm which explains intelligence\cite{citeulike:13329708, Hawkins:2004:INT:993636, kurzweil2012create, HintonSejnowski:86, domingos2015master}{}. It has been considered ever since Mountcastle's discovery of the simple uniform architecture of the cortex\cite{Mountcastle:1978}{} (six horizontal layers organized into vertical structures called cortical columns; the columns can be thought of as the basic repeating functional units of the neocortex), which might suggest a that all brain regions perform similar operations and there are no region-specific algorithms. Another famous experiment which reignited the same idea, showed, that after rewiring, the auditory part of the brain in ferrets was able to learn to interpret visual inputs\cite{roe1992visual}{}. However, alternative theories of such a structural arrangement exist, which propose that the brain might be a collection of microcircuits, that basically look the same, however perform very different functions\cite{citeulike:12886889}{}. This paper does not consider the latter view. 
Despite the fact that there is still much to be discovered, there are many facts that are either already known or are very likely to be true. Assuming that a general purpose learning procedure does indeed exist, this paper lists its key aspects that could agree with the pieces of evidence which have been gathered so far. Our knowledge about necessary ingredients of such an algorithm is shaped by neuroscientific discoveries, empirical evaluation of effectiveness of algorithms, metacognition and observations. Some of the points below may be considered as very general assumptions for reverse-engineering the learning algorithm.

\subsection{Unsupervised}
In real world, almost all data is unlabeled. Although, to the best of knowledge, nobody has discovered the precise rules used by human brain for learning, one can assume that we learn mostly in an unsupervised way. Specifically, when newborn learns about the world and how different objects interact, there might not even be a way to provide supervised signal to him/her, because the appropriate sensory representations (i.e. visual, auditory) need to be developed first.  Another piece of evidence against supervised learning may be obtained by simple calculation: assuming that there are approximately $10^4$ synapses and $10^9$ seconds of human lifetime, there is enough capacity to store all memories at the rate of $10^5$ bits/second\cite{Schmidhuber:07alt}{}, so there would simply not be enough information in the labels alone. This motivates the hypothesis of predominance of unsupervised learning, since the only way of acquiring so much information is by absorbing data from perceptual inputs\cite{hintonlecture}{}. Even when a teacher is present, most learning must be done by learning associations between events without supervision. That is to say, first learning a concepts by forming internal representations of the experiences and associating those internal representations with  names afterwards, so that the two are separated (which also allows to label all previously seen and all yet to be seen objects which fall into the same cluster). Unsupervised learning has been researched extensively and was found to be closely connected to the process of entropy-maximization, regularization and compression\cite{DBLP:journals/corr/abs-1206-5538, MacKay:itp, hinton1999unsupervised, 888}{} which means that through evolution, our brains have adapted to act as data compactors. In particular, the goal of unsupervised learning might be to find codes which disentangle input sources and describe the original information in a less redundant or interpretable way\cite{888}{} by throwing out as much data as possible out without losing information. An example of this operation has been observed in the visual cortex\cite{wiesel:1959} (but might even happen as early as in the retina) which learns patterns appearing in the natural environment and assigns high probability to those\cite{hyvarinen2009natural}{}, whereas low probability to random combinations. The real world data is said to lie near a non-linear manifold\cite{bengio2004discovering} within the higher-dimensional space, where the manifold shape is defined by the data probability distribution. Clustering is then equivalent to learning those manifolds and being able to separate them well enough for a given task.

\subsection{Hierarchical}
Humans learn concepts in a sequential order, first making sense of simple patterns and representing more complex ones in terms of those previously learned abstractions. The ability to read, might serve as an example. First we learn to see, we recognize pen strokes, then letters, words and then we are able to understand complex sentences, whereas the non-hierarchical approach would be to attempt to read straight from ink patterns on a piece of paper. The brain might have adapted this way to reflect the fact that the world is inherently hierarchical. And this observation also inspired the deep learning movement, which used the hierarchical approach to model real world data,  achieving unprecedented performance on many tasks. The way deep learning algorithms automatically compose multiple layers of representations of data gives rise to models, which yield increasingly abstract associations between concepts (hence the other names used - representation learning\cite{Barlow:89review, Bengio:2013:RLR:2498740.2498889, bengio2013deep, deng2014deep, erhan2010does} and feature learning\cite{DBLP:journals/corr/abs-1206-5538}{} among others). The main distinction between the the \emph{deep} approach and previous generation of machine learning is that the structure in the data should be discovered automatically by a general-purpose learning procedure, without the need to hand-engineer feature detectors\cite{DBLP:journals/corr/abs-1206-5538, LeCun2015}{}. This scheme agrees very well with the idea of unsupervised learning mentioned above.  In a way, abstract hierarchical representations might be a natural by-products of data compression\cite{888}{}. Upon the theoretical and empirical evidence in favor of the deep representation learning, one could formulate a requirement for any type of brain-like architecture to be deep. The question, however, is the nature of the learning procedure.



\subsection{sparse distributed representations}
\begin{figure}[!htbp]
\vskip 0.2in
\begin{center}
\centerline{\includegraphics[width=0.8\columnwidth]{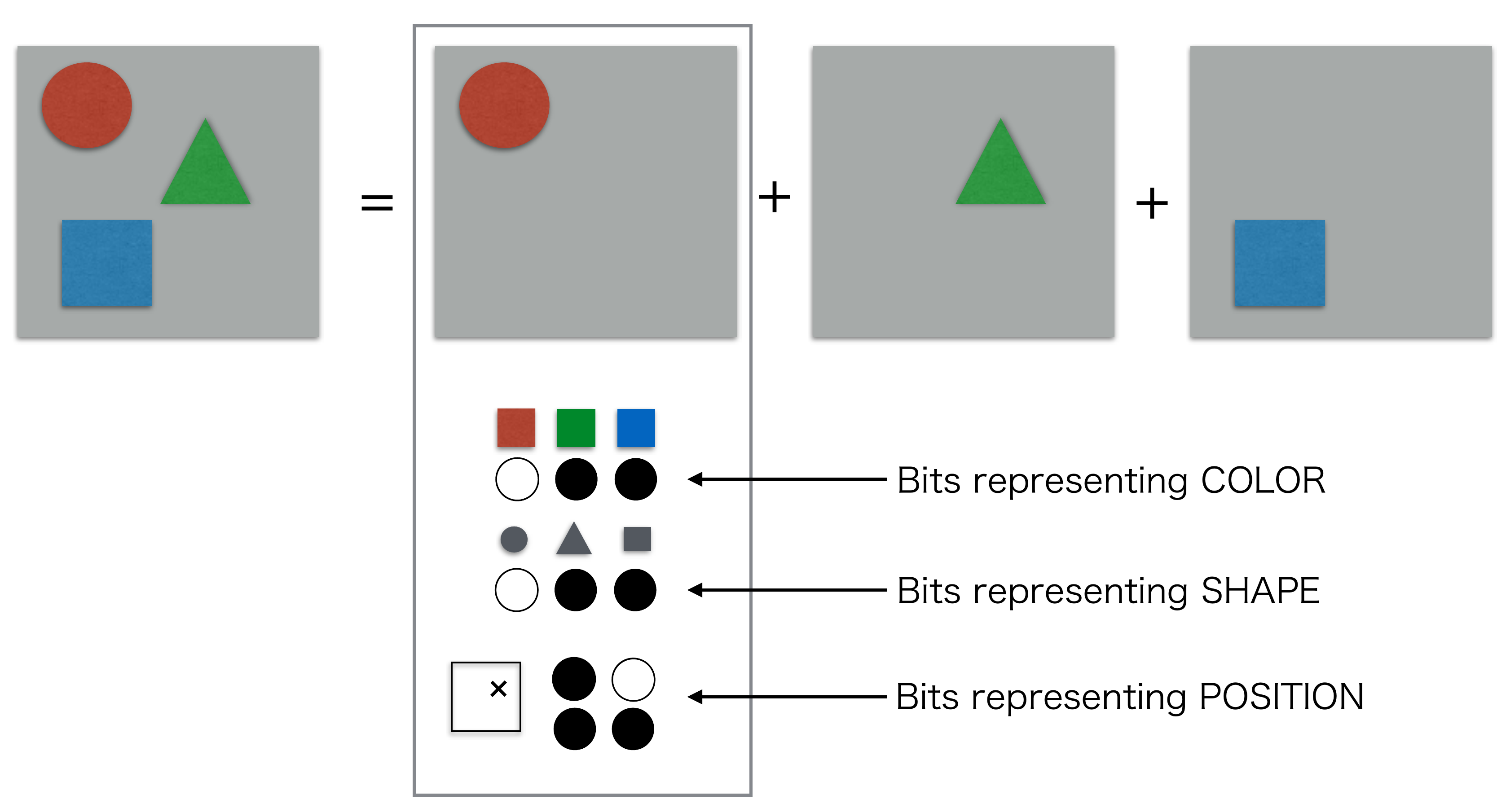}}
\caption{An example of a factored representation - where a composite object/scene can be represented as a sum of objects comprising the scene (square, circle and triangle in this case are the disentangled factors of variation). In addition to that, each of those shapes is described by reusable properties such as color or shape. }
\label{sdrs1}
\end{center}
\vskip -0.2in
\end{figure}

The existence of cortical columns in the neocortex has been linked to the functional importance of such an arrangement. Each column typically responds to a sensory stimulus representing a certain body part or region of sound or vision, so that all cells belonging to that cell are excited simultaneously, therefore acting as a feature detector. At the same time, a column which is active (receives strong input signal and spikes) will prohibit other nearby columns from becoming active. This lateral inhibition mechanism leads to sparse activity patterns. The fact that only strongly active columns will not be inhibited forces the learned patterns to be as invariant as possible, giving rise to independent \emph{feature detectors} in the cortex\cite{Bell97the`independent}{}. As it might have been expected, these sparse distributed representations in the brain (SDRs) are not coincidental, since they possess important properties from the information-theoretical perspective. The \emph{distributed} aspect makes factored codes possible, which is important in order to disentangle the underlying causes\cite{Bengio:2013:RLR:2498740.2498889} (i.e. melody, instrument, pitch, loudness, the equivalent task is called blind source separation in cocktail party problem), while sparsity affects other elements of learning good features(see Fig. \vref{sdrs1}{}). It has been proven that given certain sparsity, a signal may be correctly reconstructed even with fewer samples than the sampling theorem requires\cite{citeulike:2688127, Donoho:2006:CS:2263438.2272089}{}.

As a comparison, a dense binary representation is very space-efficient, in a sense that given $n$ bits, it is capable of storing $2^n$ different values. In the domain of learning complex patterns in real environment, there is a fundamental problem with it. Almost any random flip of bits (noise) will produce a value that is very different than the original one, so that given a noisy version of a pattern, it would not possible to recover the correct version. Sparse distributed representations (SDRs) on the other hand assume that every bit has some meaning, only a small number of bits should be active (sparse) at any moment in time and that a object is a collection of simples patterns (distributed). These properties make SDRs significantly more noise-resistant\cite{1503.07469}{}. Another important property of distributed representations is that the number of distinguishable regions scales exponentially with the number of parameters used to describe it. This is not true for non-distributed representations. That is, sparse distributed representations are combinatorially much more expressive. Given this observation, it is simple to see that from the discriminative point of view or higher levels of abstractions, SDRs will be a preferred way of representing inputs, since the the learning procedure produces a form which preserves as much information as possible while making the code as short/simple as possible (also it corresponds to finding minimum-entropy codes\cite{Barlow:89, hyvarinen2001}{}). This is in-line with the Occam's Razor or Minimum Description Length (MDL) rules which postulate that the simple solution should be chosen over the complex ones\cite{Solomonoff:64, Rissanen:78}{}. This allows for manipulating sparse representations throughout the large network and simplifies learning higher level concepts (see \emph{dimensionality reduction\cite{HinSal06, saul2003}}, \emph{redundancy reduction\cite{Li:2008:IKC:1478784, doi:10.1080/net.12.3.241.253}}).


Ever since the discovery of selective features detectors such as edge detectors and center-surround receptive fields in V1 by Hubel and Wiesel in 1959\cite{wiesel:1959}{}, learning biologically plausible sparse distributed representations of input patterns has been a hot research topic\cite{Barlow:89, Foldiak:95, Kanerva:1988:SDM:534853, Olshausen97sparsecoding}{}. It may be seen as an instantiation of unsupervised learning and many algorithms have been invented so far\cite{Hinton:1986:DR:104279.104287, ackley1985learning, Poultney06efficientlearning, 1503.07469, sparse2007ng, lee2007sparse, Hinton:97, AISTATS2011_GlorotBB11, hinton2007learning, hyvarinen2009natural, olshausen2004sparse, Donoho:2006:CS:2263438.2272089, Boureau07y.:sparse, lee:2009}{} (they include Factor models, PCA, RBM, ICA, Sparse coding, AE, among others). Convolutional Neural Networks (CNNs or convnets\cite{Fukushima:1979neocognitron, lecun-89e}{}) are on the other hand a supervised learning architectures based on the principle of learning a hierarchy of SDRs and currently provide state-of-the-art image recognition, proving discriminative value of learning such representations. Sparsity has been also link to quantifiably better performance on discriminative tasks\cite{AISTATS2011_GlorotBB11}{}, which may be explained by by the fact that sparse representations simplify optimization of an objective.

\begin{figure}[!htbp]
\vskip 0.2in
\begin{center}
\centerline{\includegraphics[width=0.9\columnwidth]{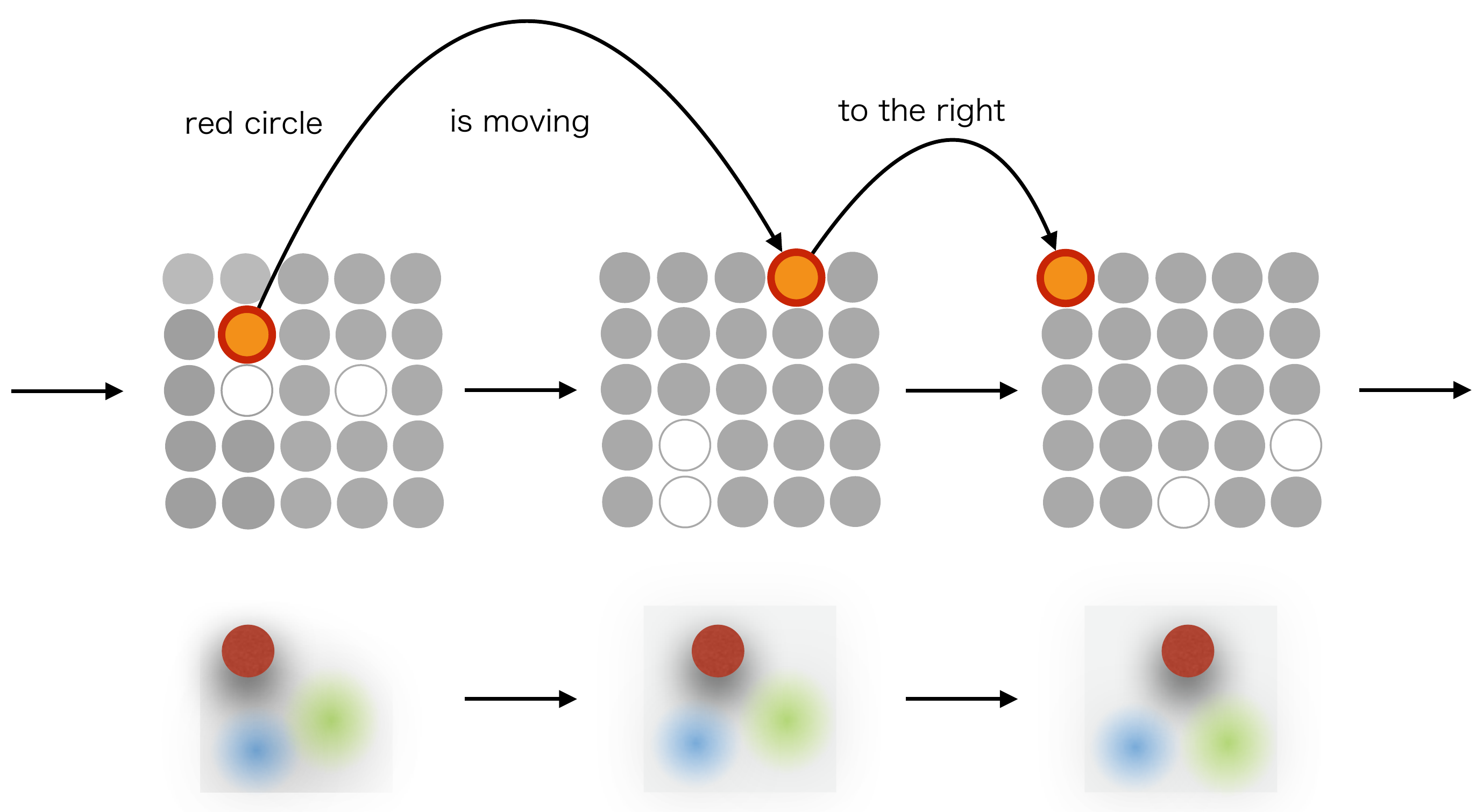}}
\caption{Efficient learning of SDRs; Sparse Distributed Representations (SDRs) simplify learning temporal dependencies; provide a mechanism for generalization and out-of-domain prediction}
\label{sdrs2}
\end{center}
\vskip -0.2in
\end{figure} 

Another desirable property resulting from a factored representation is the generalization capability, meaning that similar input patterns will produce similar bit outputs. It might imply that SDRs are a plausible candidate for the \emph{alphabet} used by the neocortex and a means to machine intelligence. One example of an advantage of an SDR compared to a dense representation becomes obvious when considering learning temporal dependencies between spatial patterns (Fig. \vref{sdrs2}). Assuming that the learning procedure has disentangled the underlying sources of variation, learning complex sequences may be decomposed into finding relationships between those sources.


%

\subsection{Objectiveless}
\label{s:objectiveless}
The backpropagation algorithm\cite{opac-b1081822,Rumelhart:1988:LRB:65669.104451}{} lies at the heart of most of deep architectures. It specifies how the internal parameters of a model at all levels should be changed in order to improve it (specifies the direction of movement in the state-space). Given certain problems\footnote{this, again, is the author's view; backpropagation is not evil; it might be a different path leading to the same goal} and following the Occam's Razor rule, this paper questions whether one really needs backpropagation to learn non-trivial concepts. Usually, the algorithm computes the derivatives of the outputs and propagates them backward, which in turn rely on having an objective function, which depends on the task definition, performance criterion and other assumptions. This is definitely a problem which causes a generality/performance tradeoff and requires some a priori knowledge about a task. Another issue is related to scalability of the procedure of propagating the error derivatives backwards from a single place in a network. Next, the standard backpropagation assumes that the objective function and intermediate activations are differentiable. 

\begin{figure}[htbp]
\vskip 0.2in
\begin{center}
\centerline{\includegraphics[width=0.9\columnwidth]{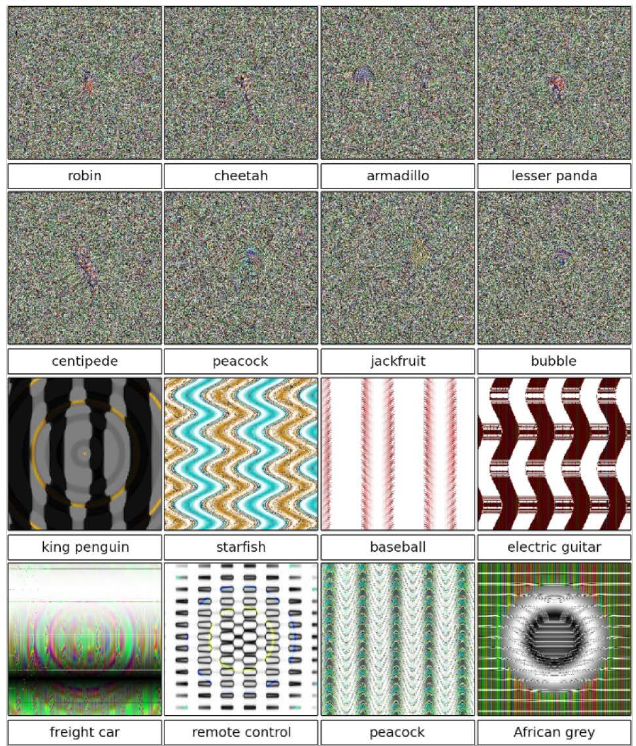}}
\caption{Images that are unrecognizable to humans, but classified by NN trained on ImageNet with with over 99.6 percent certainty. This result highlights differences between how statistical methods and humans recognize objects, figure borrowed from [\citen{DBLP:journals/corr/NguyenYC14}] }
\label{fooling}
\end{center}
\vskip -0.2in
\end{figure} 

\begin{figure}[!htbp]
\vskip 0.1in
\centering
\subfigure[]{\raisebox{0pt}{\includegraphics[width = 1.4in]{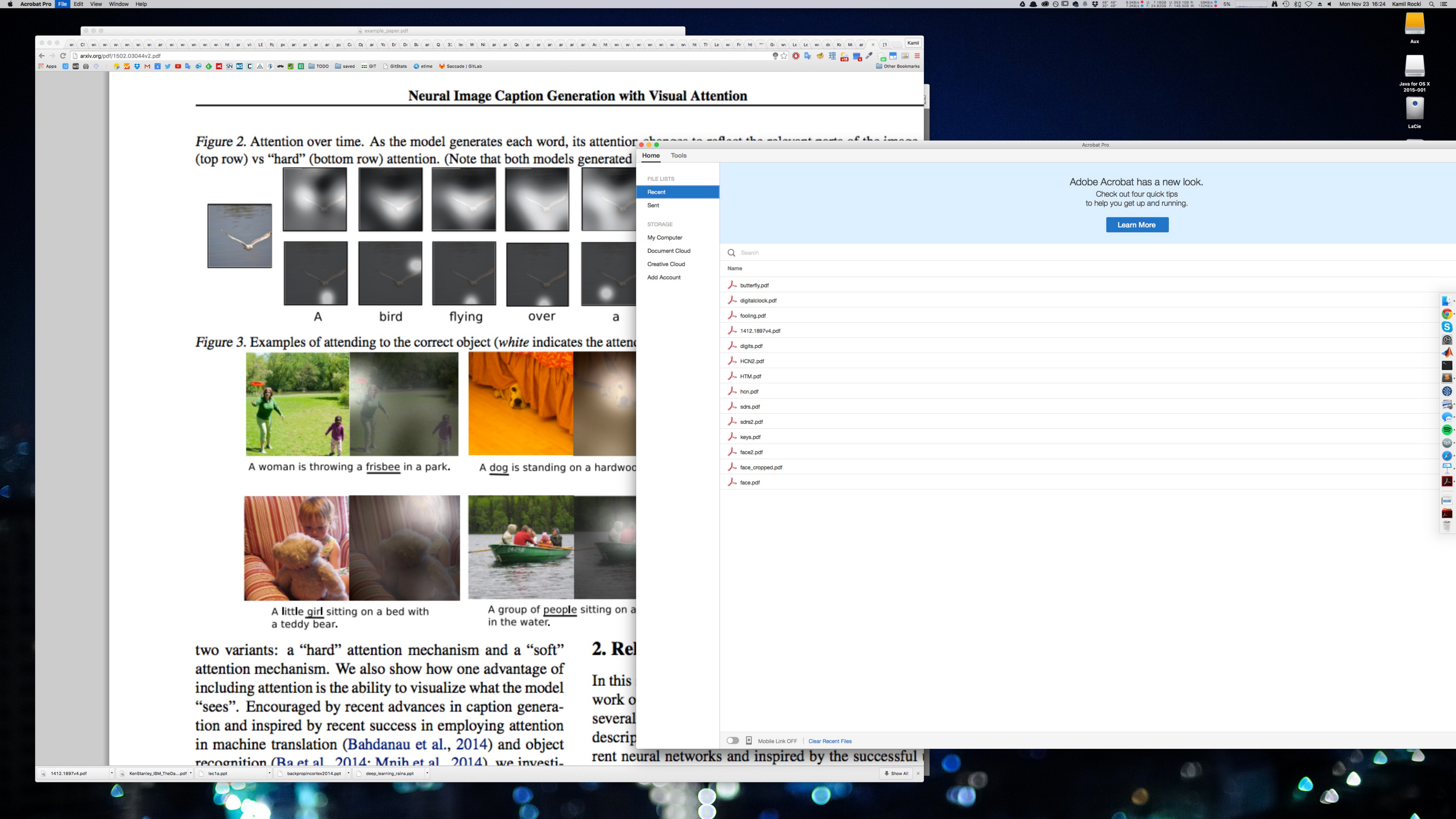}}}
\subfigure[]{\raisebox{-5pt}{\includegraphics[width = 1.4in]{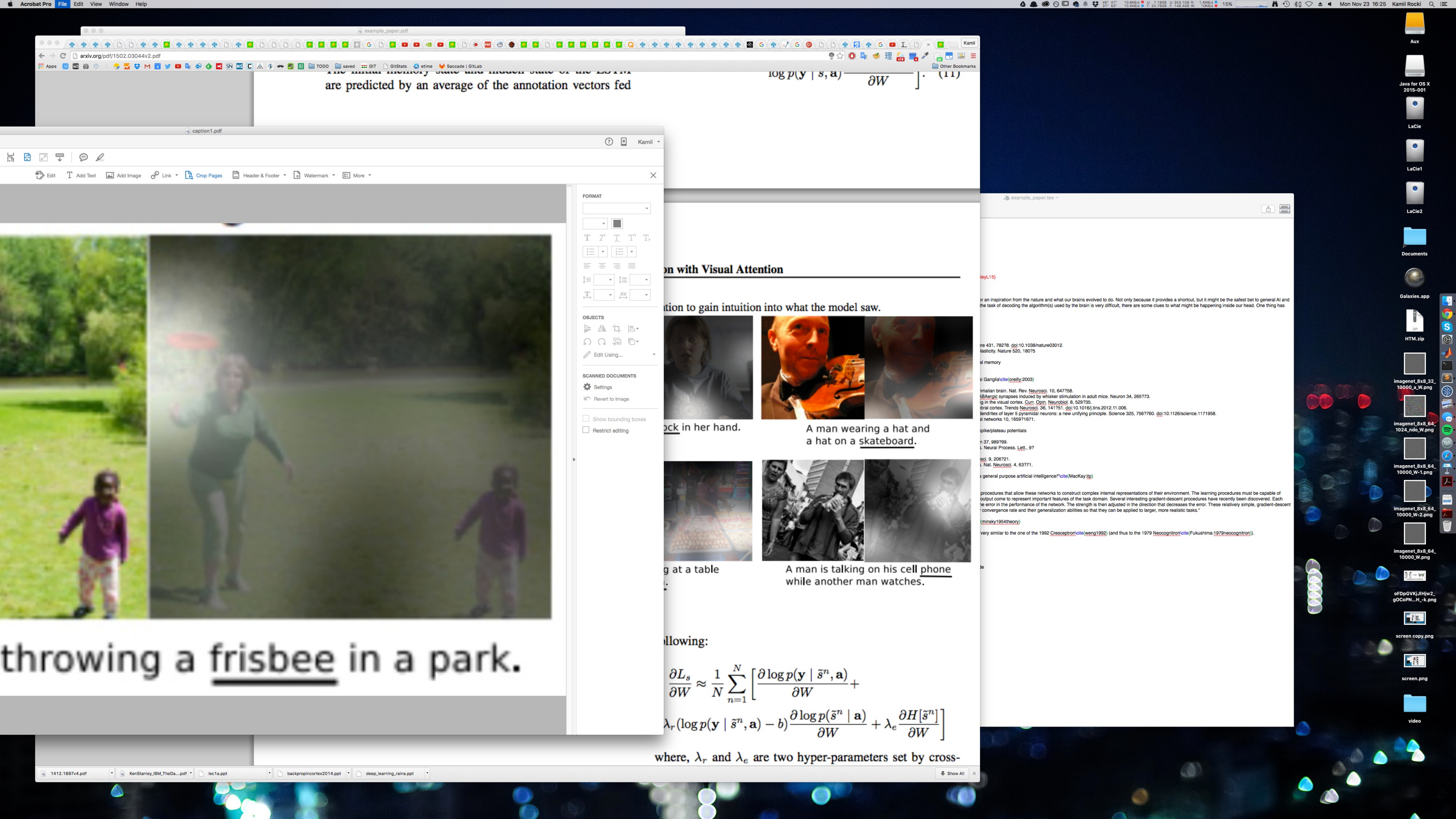}}}
\caption{Captions generated by the same network, taken from [\citen{DBLP:journals/corr/XuBKCCSZB15}]. Image (b) shows that the network actually does not have a deeper understanding of what is in an image}
\label{im_captions}
\vskip -0.1in
\end{figure}

\begin{figure}[!htbp]
\vskip 0.2in
\begin{center}
\centerline{\includegraphics[width=0.9\columnwidth]{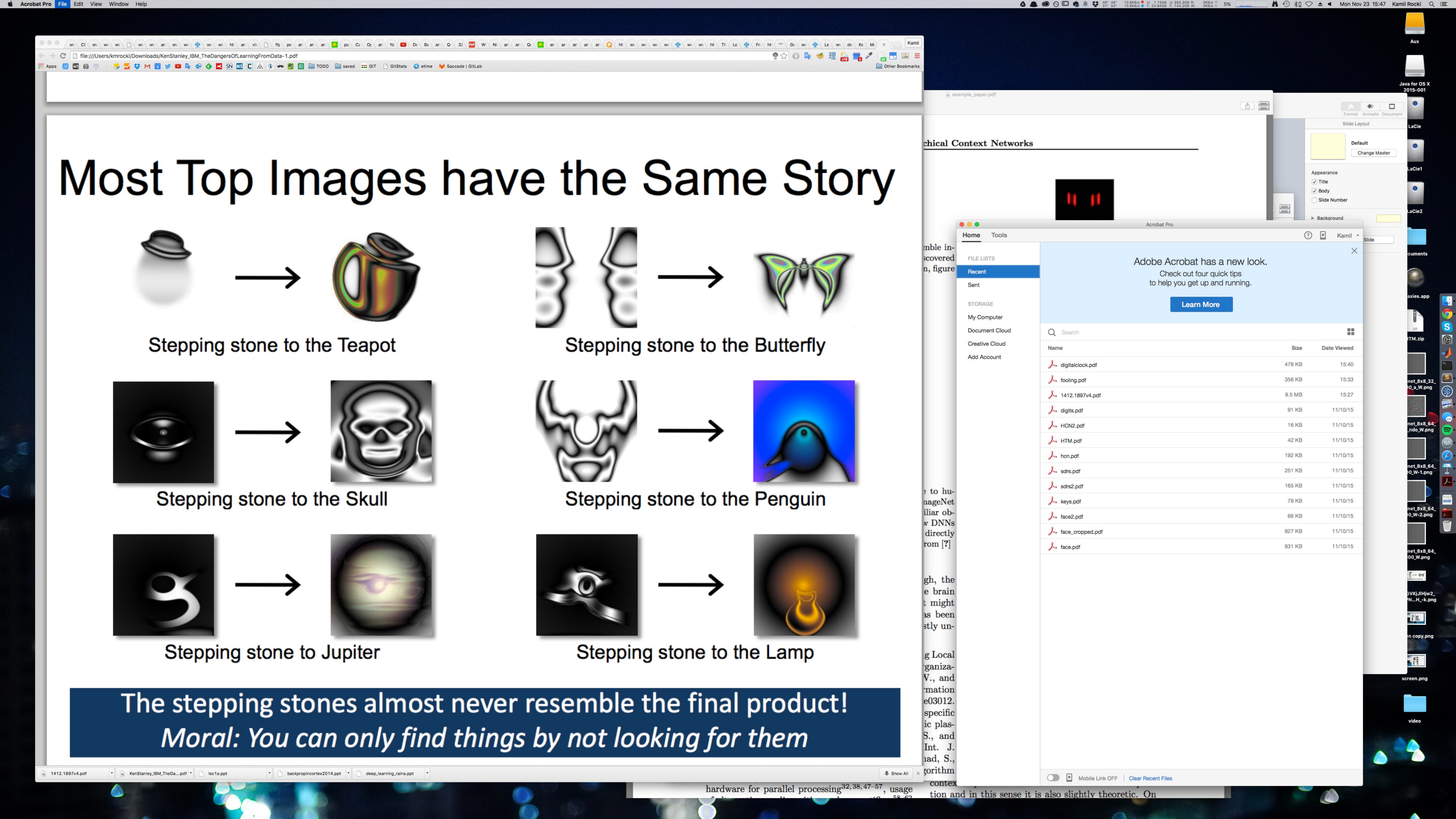}}
\caption{The final product does not need to resemble intermediate steps, some things might not be discovered when following the gradient of an objective function, figure by K. Stanley [\citen{DBLP:books/sp/StanleyL15}], see [\citen{Secretan:2011:PCS:2078014.2078016}] for other examples }
\label{butterfly}
\end{center}
\vskip -0.2in
\end{figure}

Then, in addition to the difficulty of having to hand-engineer the task definition and the objective function, there exists a problem that has been discovered recently\cite{DBLP:journals/corr/SzegedyZSBEGF13, goodfellow2014explaining} - the fact that there exist images which can be classified as almost any object with great confidence\cite{DBLP:journals/corr/NguyenYC14}{} (Figure \vref{fooling}), despite the fact that they might not resemble any know object to humans. The existence of those images suggests that the models fails to really \emph{understand} the concepts and instead the situation bears resemblance to the \emph{Chinese Room} argument\cite{searle1984minds}{}. The adversarial examples lie in the neighborhood of the data manifold, possess similar statistics, therefore the algorithm thinks they are the same, however, drastically different for humans. One hypothesis explaining this phenomenon is that having an objective\cite{DBLP:books/sp/StanleyL15}{} is the problem itself. It could be hypothesized, that by following the gradient of the objective function, one may prohibit the learning procedure from discovering the unknown state-space or that progress in learning is not equivalent with being close to the objective (Figure \vref{butterfly}). The \emph{fooling images} problem is not limited to a particular NN algorithm, architecture or dataset. In fact, it has been shown in many areas and the same undesirable properties can even be transferred from one net to the other. A striking example of contrast between how accurate NN can be at generating image captions and the type of mistakes it makes, is shown in Figure \vref{im_captions}{}. The same problem of lack of understanding of grammar and complex concepts applies to machine translation. 

Finally, one more problem which has been attributed to gradient-based learning is called \emph{catastrophic forgetting}\cite{Goodfellow2014}{}, which means that a model can \emph{forget} previously learned knowledge upon a presentation of new data by re-adjusting the parameters according to the gradients.


\subsection{Scalable}
The brain comprises approximately ${10}^{11}$ neurons and ${10}^{15}$ synapses\cite{Chklovskii2004, hawkins2015neurons}{}. In such a large network, having a single learning objective and propagating error derivates backwards might not be the best choice (one of the reasons might be the that modification of all synapses every time is wasteful, another is the hardness of parallelizing global optimization problems, however there are some approaches to solving this problem \cite{dean2012, coates:2013icml}). Instead it might be more reasonable to separate local learning (gray matter) from adjusting higher level connections between layers/regions (white matter). This functional distinction would reflect the structural hierarchy that is so predominant in \emph{deep learning} methods described before and the real world. Biological, technological, social networks (most crucially transportation) and other types of real-world networks are neither completely random nor definitely regular. Instead, their topology lies somewhere in between. The, so called, small world networks\cite{watts1998cds} may be nature's solution to a hierarchical structure which allows for separate parallel local and global updates of synapses, scalability and unsupervised learning at the lower levels with more goal-oriented fine-tuning in higher regions. 

\begin{figure}[!htbp]
\vskip 0.2in
\begin{center}
\centerline{\includegraphics[width=0.5\columnwidth]{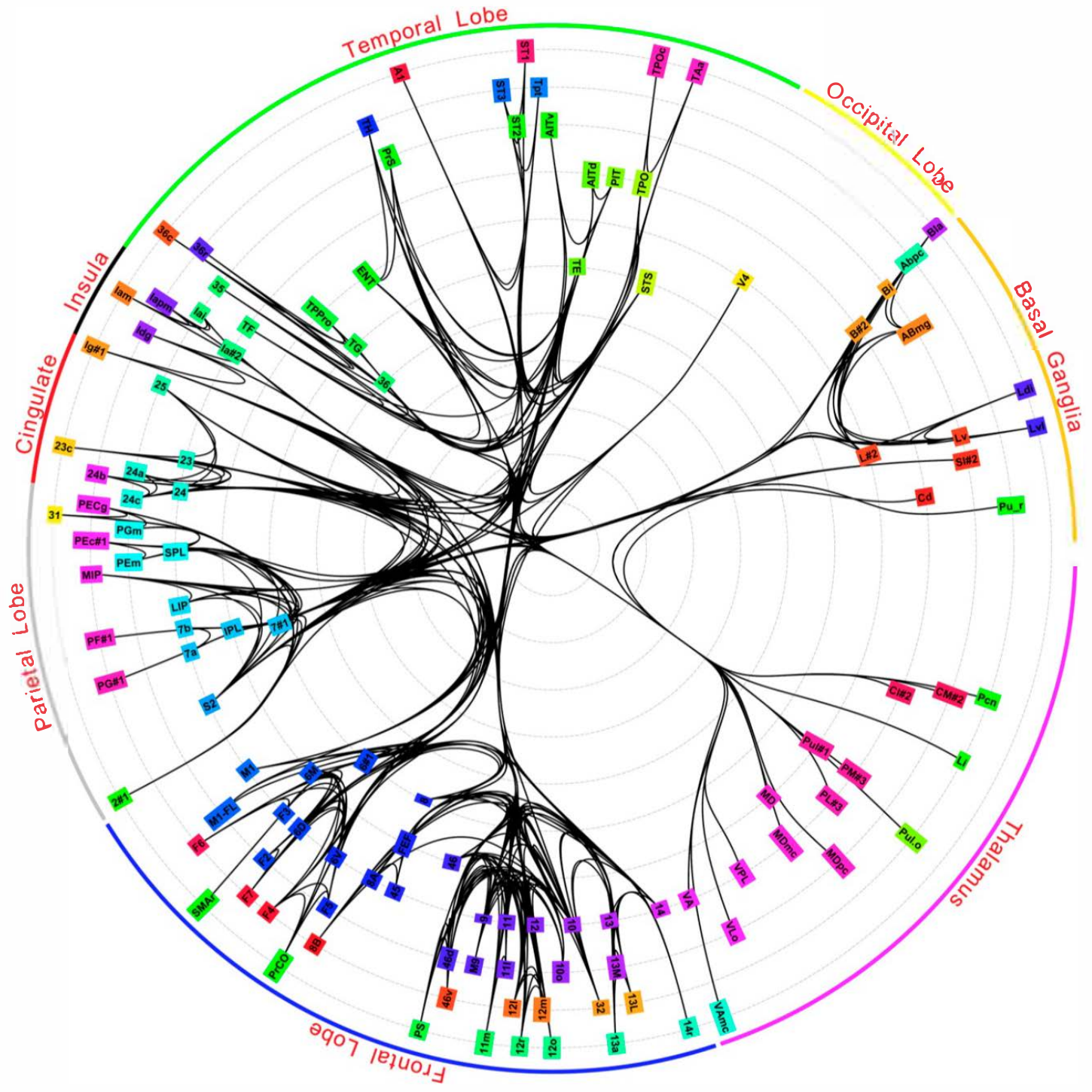}}
\caption{An example of a small world network: each edge encodes the presence of long-distance connection between corresponding regions in a macaque brain. Figure borrowed from [\citen{modha2010network}] }
\label{small}
\end{center}
\vskip -0.2in
\end{figure} 

Studying the neocortex indeed reveals that this is the case (Fig. \vref{small}), where columnar organization reflects the local connectivity of the cerebral cortex. Another piece of evidence comes from the success of convolutional networks, where sharing connections imposes such local/global structure of learning. Usage of sparse distributed representation is also very important from scalability point of view, since the representations are inherently fault tolerant. Moreover, sparse activations may be stored in a more compact only non-zero elements could be processed instead of all. Finally, it has been shown recently, that neural networks are able to learn even with very limited computational precision\cite{DBLP:journals/corr/GuptaAGN15, DBLP:journals/corr/CourbariauxBD14}{}, stochastic approximations\cite{lin2015neural} and noise\cite{Srivastava:2014:DSW:2627435.2670313, icml2013_wan13, courbariaux2015binaryconnect}{}. In fact, such networks may even have better generalization capability.

The brain is an inherently parallel machine, without a separate instruction-issuing and memory storage areas. Instead, all parts of the neocortex participate in both. This is a very big difference when compared to the von-Neumann architecture describing majority of computing systems are organized. The main bottleneck current systems concerns data movement, which implies additional bandwidth, power and latency requirements. CPUs are typically optimized for serial tasks, mitigating the negative effects of such an architecture by deep cache hierarchy, but losing when parallelism is involved. GPUs have more brain-like layout, with more \emph{equal} processing units, each having some private memory, so that they can actually operate in parallel without colliding. However, the problem of moving the data still exists, either between CPU and GPU or inside in the GPU. The same problem persists. In fact, it is quite easy to show, that it is virtually impossible to achieve the peak performance of those processors, because the data cannot be fed fast enough. Moreover, the data transfers are the major energy consumption factors on parallel GPU-like devices\cite{Villa:2014:SPW:2683593.2683684}{}. Therefore, a more radical approach may be needed in order to improve the performance significantly. The von-Neumann architecture needs to be changed into one where memory itself can compute. Some hardware which allows such a functionality has already appeared\cite{dlugosch2014efficient}{}. The concept of in-place processing assumes however, that a different approach is also needed when thinking about algorithms. This process of communication-aware algorithm design has already started with the advent of multi-core CPUs, GPUs and FPGAs. The next step is to design communication-less algorithms\cite{Baboulin201217}{}. This is an ongoing effort in supercomputing community, where it has been noticed, that no significant progress can be made without reducing information transfer-overhead.


\subsection{Biologically plausible}

%

Given the views expressed above, how could one facilitate learning? The most successful learning algorithms at the moment are based on global gradient descent algorithm, however, due to several reasons (mentioned before), the preferred solution to this problem should avoid specifying a global objective. One idea is to look once more for an inspiration in nature, and to study what our brains evolved to do in order to build rich internal representations of the environment. Not just because it provides a shortcut, but it might be the safest bet to general AI, at the same time potentially solving many other problems that we have not even though about yet. This this a major goal in current AI research. Although, the task of decoding the algorithm(s) used by the brain is very difficult, there are some clues to what might be happening inside our heads. 

To start with, very roughly the brain can be divided into separate subsystems (the exact functions are yet to be discovered), where the neocortex is the main information processing workhorse and therefore considered as being separated from lower-level actions (cerebellum), reward/value-like inputs (amygdala, limbic system) or long-term memory access/formation (hippocampal complex). This may serve as a reason why this paper skips those parts, and may justify why it is not unreasonable to think about a single learning algorithm as not being reinforcement-based.

It has been shown that two of the tasks being performed in the visual cortex in the brain is spatial pattern detection\cite{wiesel:1959}{} and forming sparse representations of the inputs\cite{Olshausen97sparsecoding}{} (introduced earlier), which have been found to be work in the same way when modeled algorithmically too\cite{lee2007sparse, Le_buildinghigh-level}{}. It has been shown that biologically plausible features can be learned using very simple Hebbian-like learning rules\cite{Linsker1986}{}. Then, it has been observed that there exist so called \emph{simple} cells, which act as a specific pattern detector, such as an oriented edge, and \emph{complex} cells, which are to some extent invariant to transformations of the inputs and react to a more general group of stimuli (i.e. shape detector). Those discoveries served as an inspiration to the groundbreaking performance of deep convolutional neural networks in image recognition\cite{Fukushima:1979neocognitron, lecun-89e, lee:2009, NIPS2012_4824}{}. At the same time, those successes served as a feedback loop to study create even more accurate models of processing in the brain, such as a feedforward HMAX\cite{Riesenhuber:99} model, whose neurophysiologically plausible topology is very similar to the one of the 1992 Cresceptron\cite{weng1992} (and thus to the 1979 Neocognitron\cite{Fukushima:1979neocognitron}). Similarities have been in the auditory cortex\cite{4218213}{}, where individual phonemes activated different subsets of auditory neurons.

When it comes to learning, there is still much to be discovered. One thing has been already pointed out, the algorithm/learning process has to be be mostly unsupervised in nature. More specifically, it has been shown and hypothesized that the main function of the brain has to be unsupervised learning of temporal sequences. At a very general level it means that is constantly anticipates an outcome, acts, observes the world, compares observations with previous expectations and adjusts (or forms new) synapses so that the internal model of the world makes more accurate predictions\cite{Hawkins:2004:INT:993636}{}. More formal approaches expressing the same idea have been formulated by G. Hinton (Boltzmann machines\cite{HintonSejnowski:86}) and K. Fritz (free-energy principle\cite{Friston2010}), Wiskott and Sejnowski (slow feature analysis\cite{WisSej2002}). Local contrastive divergence-like (CD) learning or target propagation\cite{DBLP:journals/corr/LeeZBB14} may be a much more plausible method than backpropagation\cite{hinton2007backpropagation}; On the implementation side of things, it has been discovered that there are at least 3 types of connections being integrated in a pyramidal cell\cite{hawkins2015neurons,Lamme1998529} (proximal, distal/basal, apical) most likely serving feed-forward, sequence and feedback roles accordingly using simple, local Hebbian-like\cite{Hebb:49, MacKay:itp} learning rules, which might resemble local backpropagation algorithm at a very tiny scale. In fact that is what has been  observed in the brain\cite{Cichon2015,Fu2011}{}. 

\begin{figure}[!htbp]
\vskip 0.2in
\begin{center}
\centerline{\includegraphics[width=0.4\columnwidth]{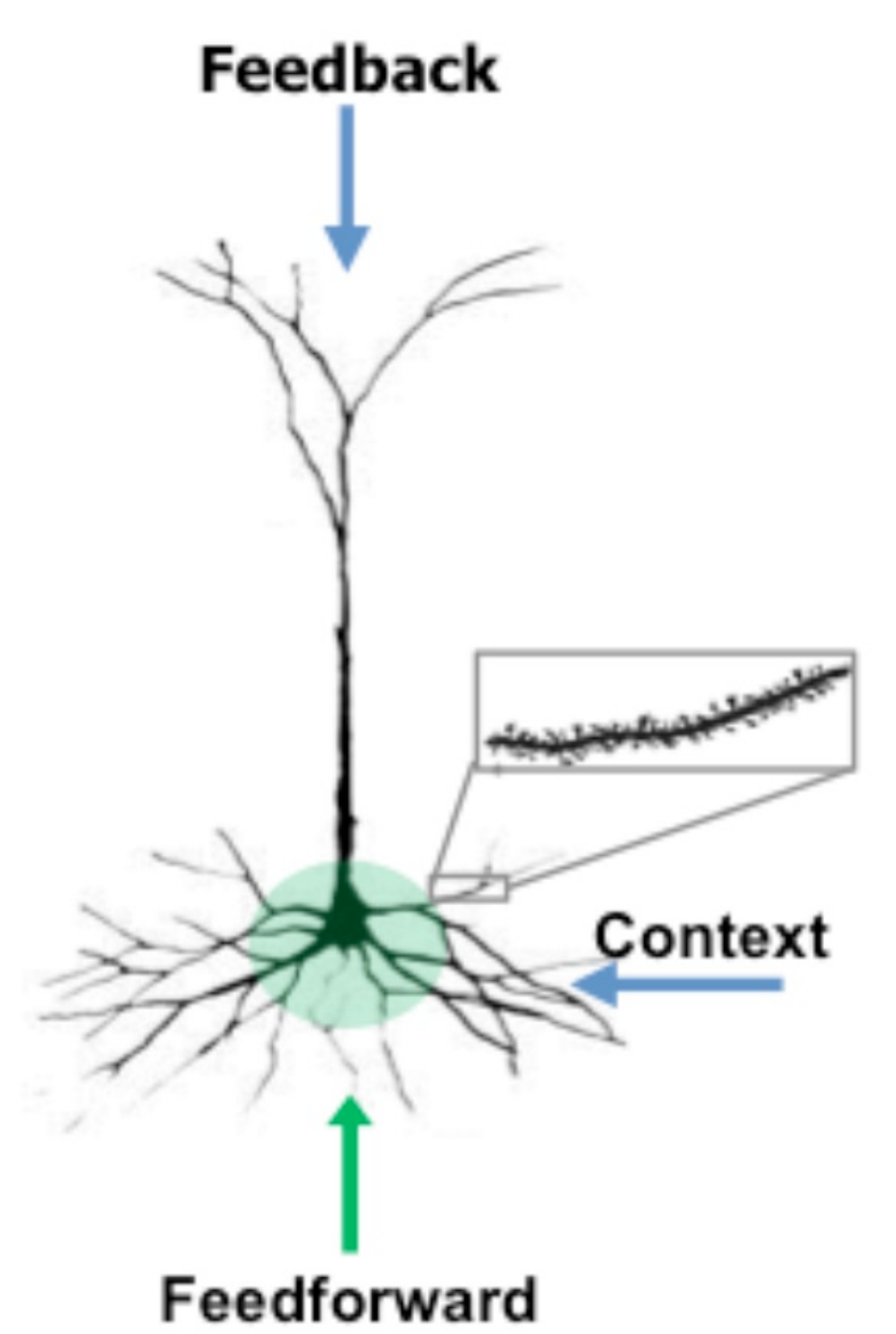}}
\caption{A neocortical pyramidal neuron has thousands of excitatory synapses located on dendrites (inset). There are three sources of input to the cell. The feedforward inputs (shown in green) which form synapses proximal to the soma, directly lead to action potentials. NMDA spikes generated in the more distal basal and apical dendrites depolarize the soma but typically not sufficiently to generate a somatic action potential. Figure borrowed from [\citen{hawkins2015neurons}]}
\label{neuron}
\end{center}
\vskip -0.2in
\end{figure} 

 

It is hypothesized that pyramidal cells in the neocortex might not integrate incoming bottom-up and top-down signals in a simple way\cite{Larkum2013141, Larkum07082009}{}, instead their operation may be more gate-like, with proximal dendrites responsible for feed-forward stimuli, distal dendrites for immediate prediction, and apical dendrites acting as a filter/gate and disambiguation mechanism\cite{Hawkins:2004:INT:993636} (logical AND-like, so creating an OR-AND or SUM-MULTIPLY cascade\cite{HäusserMel2003, representingrelations}{}). in this sense, recently very popular in ML approach of including gating as a means of memory access control\cite{DBLP:journals/corr/GravesWD14, DBLP:journals/corr/WestonCB14} or attention\cite{DBLP:journals/corr/ChorowskiBSCB15, DBLP:journals/corr/XuBKCCSZB15, mnih2014recurrent}{}, might be biologically plausible\cite{oreilly:2003, 888}{}.


Probably the most anatomically accurate model so far is the networks of spiking neurons\cite{Maas:1997:NSN:281543.281637}{}. Learning is such networks can also be some using local plasticity rules, i.e. Spike Timing Dependent Plasticity\cite{markram1997regulation} (STDP), a temporal form of Hebbian-like learning based on temporal correlations between the spikes of pre- and postsynaptic neurons. It is believed to underlie learning and information storage in the biological neural networks. It might suggest that in fact, local learning rules do exist in the brain, however in such an asynchronous, asymmetric version. The core operation which can be defined in this kind of networks is detecting the occurrence of temporally close but spatially distributed input signals\cite{ruf1997learning}{}, that is, coincidence detection. For the sake of simplicity\footnote{and to stay within a reasonable number of pages}, this paper will not focus on this aspect (However, some approaches to AI are based entirely on the fact of asynchronously propagated spikes\cite{Merolla08082014}{}; also {\vref{itm:neuromorphic}}), but assume a simpler, symmetric local learning approach as described by G. Hinton\cite{hinton2007backpropagation}{}.

An important fact about the neocortex is that connections between neurons can be created and removed, in other words, the number of parameters of the model is not fixed. However, majority of current ML approaches (deep or not) focus only on changes in the strengths of connections between neurons (weights/parameters), meaning that the topology of the network is fixed and reflects some prior knowledge. Given sparse connectivity (approx. $10^4$ synapses/neuron), an approach which is closer to the biological reality (wiring plasticity, where connections can be formed and removed\cite{Chklovskii2004}) would offer a great advantage in terms of increase of the number of combinations of neurons/synapses available to encode learned information. 




\section{Towards Machine Intelligence}



The most interesting aspect of this research is connecting the mechanisms described above with the theoretical concepts of what machine intelligence should be. Given some low-level properties of the learning algorithm, what would be the overall goal of learning and what should the learning path look like? What kind of behavior would be considered as a stepping stone towards machine intelligence and if so, is there a way to describe it in a precise way? The very basic question of what it means for a machine or an algorithm be intelligent needs clarification. According to some, goal-directed behavior is considered the essence of intelligence\cite{Russell:2003:AIM:773294}{}. However, this implies that the necessary and sufficient condition of intelligent behavior is rationality and this paper questions this statement. Humans are often very far from being rational. Creativity does not fall into this definition, risk-taking might not be rational, yet it's essential for innovation. Therefore, far more appealing theories of universal intelligence are those with broader priors, such as the theory of curiosity, creativity and beauty described by J. Schmidhuber\cite{Schmidhuber:07alt}{}.
Previous section introduces problems which may arise from objective based learning, that is the \emph{Chinese Room} argument, when all the algorithm is interested to do is to map inputs to outputs without any motivation to learn anything beyond the task given. An intelligent algorithm (strong AI\cite{searle1984minds}{}, among other names) should be able to reveal hidden knowledge which might not even be discoverable to humans. Despite not having a specific task, this section will point out functional ingredients of a any learning procedure which would not violate the generality assumption.
 
 \subsection{Compression}
Learning may be likened to a formal information-theory based concept of information compression. Assuming that the goal is to build more compact and more useful representations of the environment(such as finding minimum entropy codes\cite{journals/neco/BarlowKM89}{}), this interpretation relates to representation learning and analogy building compression scheme\cite{DBLP:journals/corr/abs-1108-1169} of the neocortex. One way of looking at this task is considering a general artificial intelligence as a general purpose compressor, one which is able to discover the probability distribution of any source\cite{MacKay:itp}{}. However, the \emph{No Free Lunch Theorem}\cite{Wolpert:1997:NFL:2221336.2221408} states that no completely general-purpose learning algorithm can exist or in other words that for every compressor, there exists a data distribution on which it will perform poorly. This implies that there must exist some restrictions on the class of problems it will work on well. The previous section already mentioned a few of them, which are fortunately very general and plausible such as the \emph{smoothness prior} or \emph{depth prior}{} (also see [\citen{Bengio+chapter2007}] for a more complete list of sensible assumptions).

\subsection{Prediction}
Whereas smoothness prior may be considered as a spatial coherence, the assumption that the world is mostly predictable corresponds to temporal or more generally spatiotemporal coherence. This is probably the most important ingredient of a general-purpose learning procedure. In other words, it states that things which close in time are close in space and vice versa. A purely spatial analogy is huge image space and only a tiny fraction of possible real images\cite{hyvarinen2009natural}{}. The same is true for spatiotemporal patterns; the assumption that a  sequence of spatial patterns is coherent, restricts the spectrum of future spatial states which are likely (if looking at a giraffe, the next thing you see is most likely a giraffe too, images of trains could be probably excluded from predictions).

Occam's Razor rule or MDL principle\cite{Solomonoff:64, Rissanen:78}{} state that simple solutions should be favored over more complex ones, therefore learning better representations should be a goal itself, even without any other objective. If it is assumed that no task is given a priori, the best we can do is just to observe and learn to predict. One of the first working examples (and a proof of concept) is the principle of history compression employed in the recurrent architecture proposed by J. Schidmuber\cite{schmidhuber1992}{}.

\subsection{Understanding}
The ability to predict is equivalent to understanding, since at any given moment, a cause and prediction could be inferred from given state context. Therefore, learning to predict may be a more general requirement of an intelligent behavior. In fact is has been postulated\cite{Hawkins:2004:INT:993636}{}, that all the brain does is constantly predicting the future states, comparing those predictions with sensory inputs and readjusting accordingly. This might seem to be equivalent to backpropagating the error through the entire network, however from the biological perspective the prediction/expectation readjustment of neurons is most likely operating locally.

\subsection{Sensorimotor}



\begin{figure}[!htbp]
\vskip 0.2in
\begin{center}
\centerline{\includegraphics[width=\columnwidth]{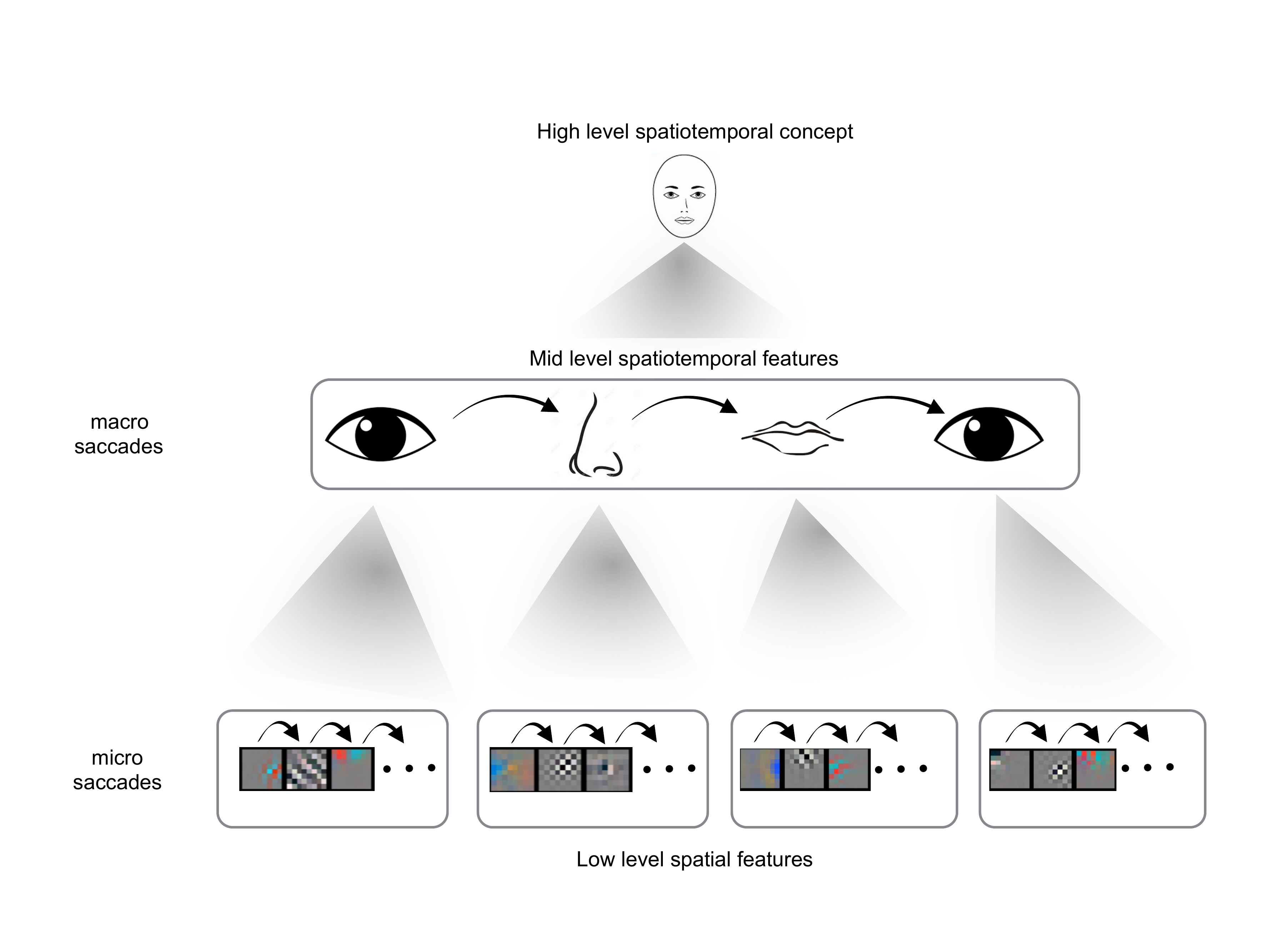}}
\caption{Face as an example of a spatiotemporal concept, micro-saccades are sequences of low-level spatial patterns in the fovea, they can be pooled temporally into a mid-level concept of an eye, or nose; macro-saccades are more task-oriented movement - moving between nose, eyes, mouth}
\label{face}
\end{center}
\vskip -0.2in
\end{figure}

\begin{figure}[!htbp]
\vskip 0.2in
\begin{center}
\centerline{\includegraphics[width=2.5in]{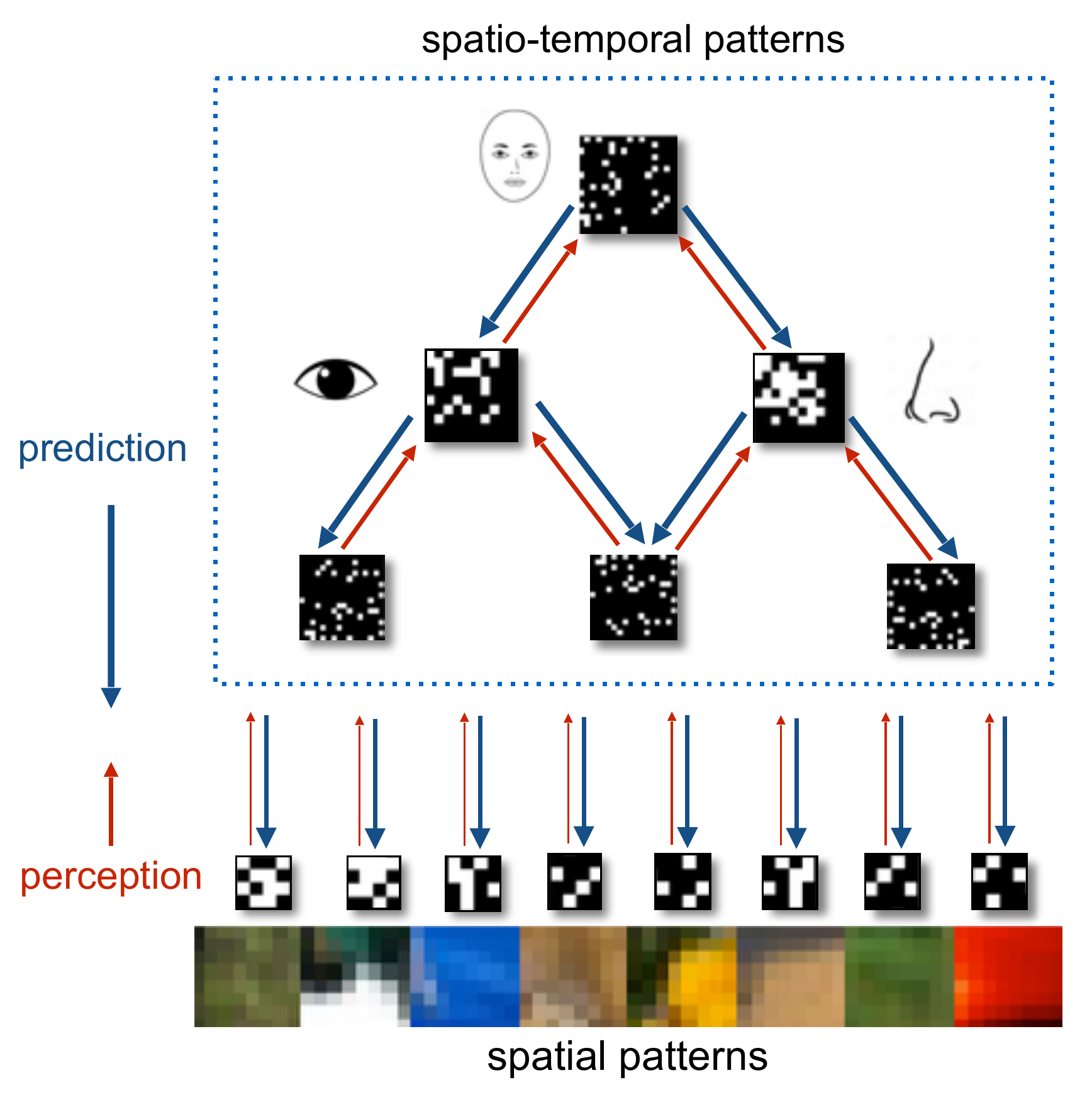}}
\caption{Feedforward and feedback connections' roles in concept understanding. Despite rapidly changing sensory inputs and different order of observations, the brain is capable of maintaining stable representations at higher levels; low-level predictions depends on context (eye, mouth, face)}
\label{face2}
\end{center}
\vskip -0.2in
\end{figure}

Scientists have demonstrated that the brain predicts consequences of our eye movements based on what we see next. The findings have implications for understanding human attention and applications to robotics. Despite the fact that, in practice, the is no experienced is perceived twice, human brains are able to form a stable representation of an abstract make accurate predictions despite changes in context. An example of such mental representations being present may be observed by explaining rapid eye movements known as saccades (see Fig. \vref{face}). Our eyes move rapidly approximately three times a second in order to capture new visual information. With each jump a new image falls onto the retina. However, we do not experience this quickly-changing sequence of images, instead, we see a stable image. The brain uses such a mechanism in order to redirect attention, since only approximately 1$^{\circ}$ of the retina provides sharp image (fovea). This operation has been extensively researched from the neuroscientific perspective as it provides one of few visible brain activities\cite{Rolfs2011, Kowler20111457} as well as provided an inspiration for algorithms mimicking this behavior\cite{Gaskett:04, DBLP:journals/corr/Ranzato14, SchmidhuberHuber:91, NIPS2010_4089}. Sensorimotor connections are needed in order to know, which changes in the image do not result from internal eye movement. It is assumed\cite{yuwei2015maintaining} that the motor command are being subtracted from the inputs in order to provide an invariant representation of a concept. What this implies is that actually every part of the neocortex must be performing this function given the uniformity. One hypothesis is that the basic repeating functional unit of the neocortex is a sensorimotor model\cite{hawkins2015neurons}{}, that is every part of the brain performs both sensory and motor processing to some extent. Complex cells in V2 visual cortex are invariant to small changes of inputs patterns\cite{lee2007sparse}{}, those invariant activations might be mapped purely spatially or may represent a spatiotemporal patterns (i.e. invariant representation given an action). Other experiments support the claim, showing a similar mechanism operating on different type of sensory inputs\cite{Krieger:2015:SIW:2838985, Diamond2008}. From an implementation perspective, sensorimotor integration may be understood in the same way as top-down connections mentioned in the previous section (see Fig. \vref{face2}). 
 
\subsection{spatiotemporal concept}
This section proposes that the sensorimotor integration may be indeed a more general way the brain operates. 

Thinking about motor command in a more abstract way, it is possible to show that in order to disambiguate multiple predictions. one needs to \emph{inject additional context} as in Fig. \vref{face2}. This paper assumes that predictions are associated with some uncertainty\cite{ref1, 10.1371/journal.pcbi.1004305} as in the bayesian approach and that instead of assuming a single point prediction, the distribution is highly multimodal. Additional context is equivalent to integrating evidence which makes predictions more specific.


\begin{figure}[!htbp]
\vskip 0.2in
\begin{center}
\centerline{\includegraphics[width=0.7\columnwidth]{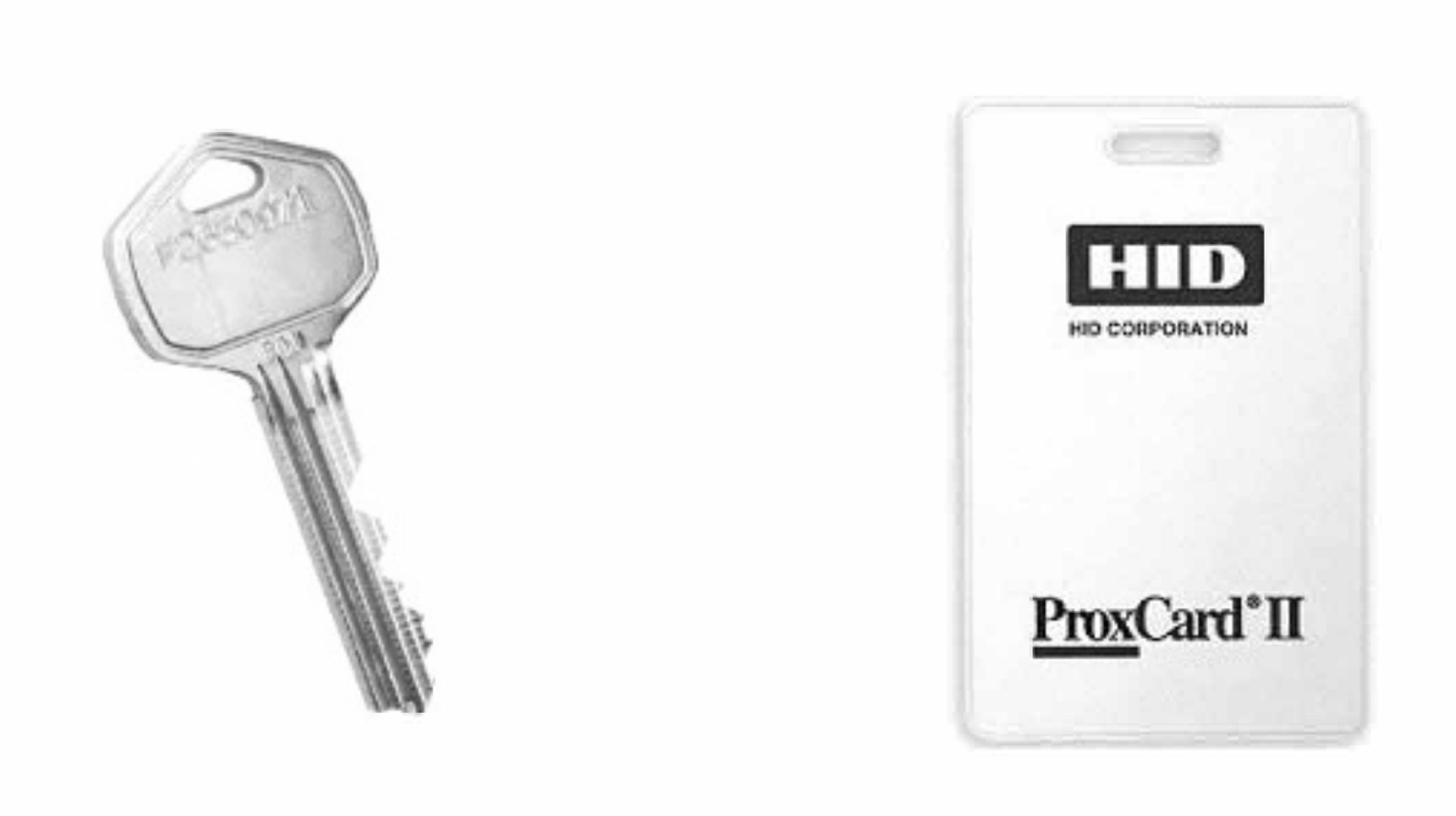}}
\caption{An example of a spatiotemporal concept}
\label{keys}
\end{center}
\vskip -0.2in
\end{figure}  
The necessity for a means of manipulating a spatiotemporal concept can be illustrated with a simple example. Given 2 images as in Fig. \vref{keys} it is obvious how unnatural is classification based on purely spatial aspect of a pattern. A much more natural way of putting these 2 objects in the same category is by their function, which requires an ability to \emph{imagine} whether a particular object can be used to be used in a certain way (in this case, open a door). The same applies to other objects, such as chairs. It is much more natural to learn these concepts as spatiotemporal ideas rather than predominant purely spatial machine-learning methods (CNNs). When considering the ability to \emph{imagine/dream/hallucinate}, then the commonness sensorimotor functionality in the brain is not very surprising. The concept of manipulating a compact spatiotemporal thought might be necessary from the reasoning perspective\cite{bottou-mlj-2013}{} or transfer learning, as majority of the analogies we make are temporal in nature. The importance of learning transformations in the real-world has been recognized in the research community\cite{memisevic2010, Boulanger-et-al-ICML2012, Sutskever, sutskever2008, WisSej2002, Elman90findingstructure, 888}{}, but still needs more attention.
\vskip 0.2in

\subsection{Context update}
\begin{figure}[!htbp]
\vskip 0.2in
\begin{center}
\centerline{\includegraphics[width=0.8\columnwidth]{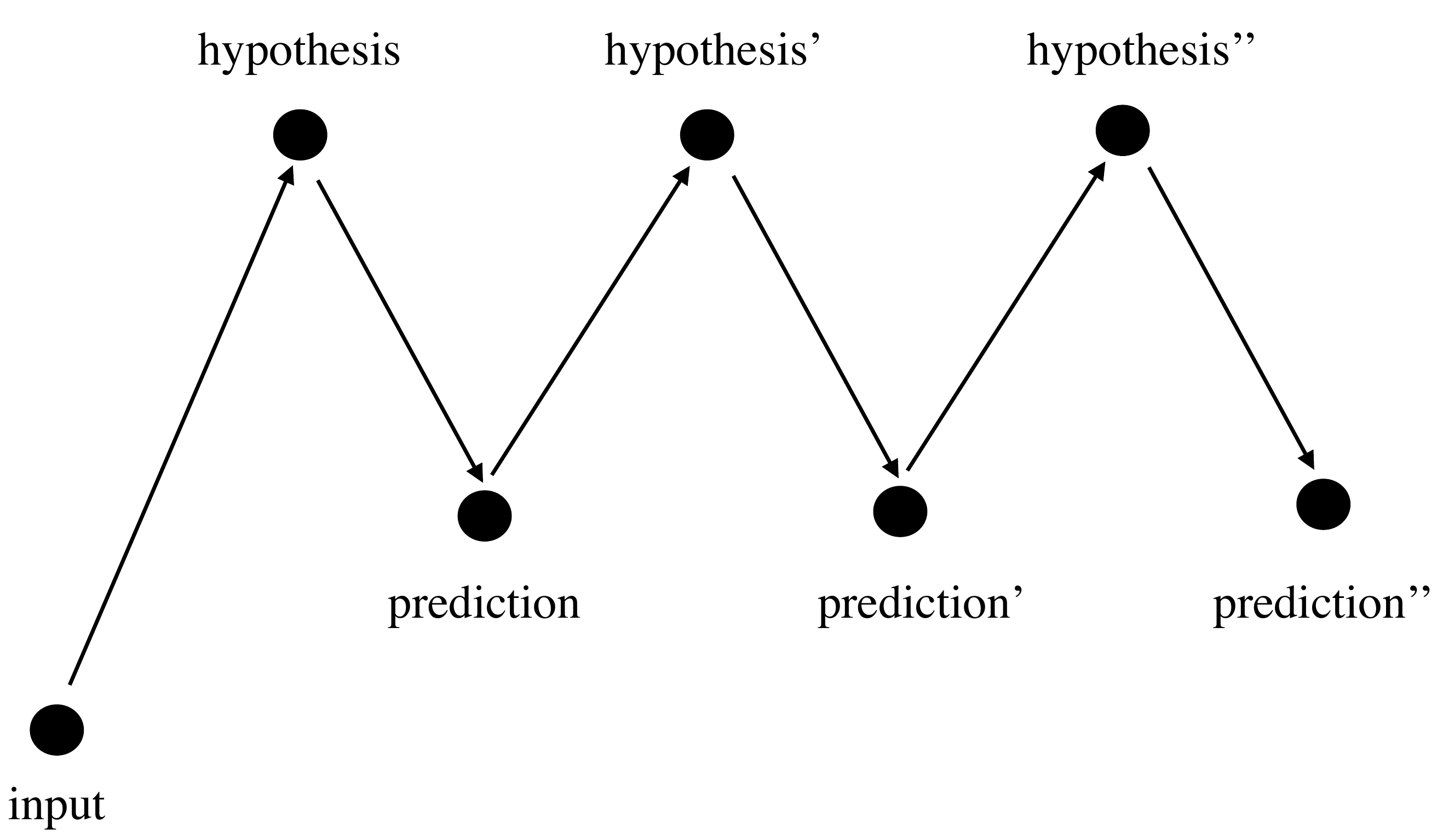}}
\caption{Illustration of iterative context update, every prediction changes the context slightly and vice-versa}
\label{f:rerun}
\end{center}
\vskip -0.2in
\end{figure}

The last functional component postulated by this paper states that there exists an infinite (in theory) loop between bottom-up predictions and top-down context. The hypothesis is that such interconnectedness enables perceptual filling in, where higher layers make hypotheses about the inferences coming from the lower layers and the predictions are iteratively refined based on those hypotheses. It may be likened to working memory theory, where non-episodic memories are being held (not involving hippocampus). An analogy of this an Expectation Maximization or the learning procedure commonly used in Boltzmann Machines, where a samples are obtained iteratively by alternating between unit activations on 2 connected layers\cite{series/lncs/Hinton12, resnik2010gibbs}{} (see Fig. \vref{f:rerun}). A real-world analogy of this process is solving a crossword or a sudoku puzzle or filling in missing words in a sentence. Those problems may require iterative procedure of refining the solution with using intermediate hypotheses. In addition, it links the ideas of objective-less learning, imagination and novel pattern discovery. 

\section*{Acknowledgments} 
Partial support for this work was provided by the Defense Advanced Research Projects Agency (DARPA). I would like to thank members of machine intelligence group at IBM Research and Numenta for their suggestions and many interesting discussions.

\section*{Remarks} 
This paper is not fixed. Its content is subject to iterative local optimization. Any comments or suggestions are welcome.

\bibliography{example_paper}
\bibliographystyle{unsrt}

\end{document}